\documentclass[11pt,a4paper]{article}
\usepackage{float}
\usepackage{url}
\usepackage{amsmath,amsfonts}
\usepackage{algorithmic}
\usepackage{array}
\usepackage{graphicx}
\usepackage[hidelinks]{hyperref}

\title{\bfseries
Guiding Sparse Neural Networks with Neurobiological Principles to Elicit Biologically Plausible Representations
}

\author{
Patrick Inoue$^{*\dag}$ \and
Florian Röhrbein$^{\dag}$ \and
Andreas Knoblauch$^{*}$ \\[1ex]
{\small $^{*}$KEIM Institute, Albstadt-Sigmaringen University, Germany}\\
{\small $^{\dag}$Department of Computer Science, Chemnitz University of Technology, Germany}
}

\begin{document}

\maketitle

\begin{abstract}

While deep neural networks (DNNs) have achieved remarkable performance in tasks such as image recognition, they often struggle with generalization, learning from few examples, and continuous adaptation—abilities inherent in biological neural systems. These challenges arise due to DNNs' failure to emulate the efficient, adaptive learning mechanisms of biological networks. To address these issues, we explore the integration of neurobiological inspired assumptions in neural network learning. This study introduces a biologically inspired learning rule that naturally integrates neurobiological principles, including sparsity, lognormal weight distributions, and adherence to Dale's law, without requiring explicit enforcement. By aligning with these core neurobiological principles, our model enhances robustness against adversarial attacks and demonstrates superior generalization, particularly in few-shot learning scenarios. Notably, integrating these constraints leads to the emergence of biologically plausible neural representations, underscoring the efficacy of incorporating neurobiological assumptions into neural network design. Preliminary results suggest that this approach could extend from feature-specific to task-specific encoding, potentially offering insights into neural resource allocation for complex tasks.

\end{abstract}

\section{Introduction}
\label{introduction}

\noindent Deep neural networks (DNNs) have revolutionized fields such as computer vision and natural language processing, achieving state-of-the-art performance across a wide range of supervised learning tasks \cite{lecun2023}. The prevailing algorithm for training DNNs is backpropagation (BP) \cite{rumelhart2}. It has played a crucial role in the advancement of machine learning \cite{balduzzi}, even surpassing human-level performance in some individual specific tasks such as image recognition and mastering complex games like Go \cite{b22}. However, while BP excels at these narrowly defined tasks, it fails to replicate the efficient learning and adaptability observed in biological neural networks. In particular, it fails to emulate the generalization, learning from few examples, and continuous learning through trial and error inherent in biological systems \cite{b22}.

These limitations may arise from several foundational assumptions that are required for efficient training using BP, yet often not subjected to sufficient scrutiny \cite{ijcnn}. These include fully connected architectures, symmetric synaptic weights for both forward and backward signal propagation—referred to as the 'weight transport problem'—and the rigidity of synaptic weights. In contrast, biological neural networks decouple the processes of signal propagation and error feedback, ensuring that error signals do not directly influence forward information processing \cite{bartunov2018}.

Addressing these and other challenges requires the development of biologically inspired alternatives that better align with the adaptive and efficient learning mechanisms observed in natural neural systems \cite{whittingtonBogacz, backprob_brain, Crick_F, Grossberg, stork1989}. Consequently, a number of such algorithms have been developed, with two principal approaches emerging. The first set of approaches, such as those by \cite{scellier}, \cite{sacramento}, \cite{lee}, and \cite{5}, aims to enhance the biological plausibility of BP, while still requiring some form of backward signal propagation, either through the error signal or surrogate methods like target or equilibrium propagation. Consequently, these approaches do not solve the weight transport problem. Another notable approach within this category is the Weight Alignment (WA) approach, proposed by \cite{5, review}, which directly addresses the weight transport problem by using distinct, arbitrary weights for BP. However, the authors themselves acknowledge, that this method fails to align with biological feedback mechanisms \cite{bartunov2018}. Furthermore, this approach still relies on error feedback that does not directly influence neural activity, as the error signals in WA are internal attributes of neurons backpropagated through feedback weights. In contrast, biological neural networks convey information through output spikes transmitted via axons and synapses, with other internal attributes of neurons remaining mostly local \cite{stork1989, review, song2020}.

The second class of approaches completely eliminates the need for a backward path, using only bottom-up information about the activities of neurons. This strategy circumvents the weight-transport problem and aligns more closely with Hebbian plasticity theories, where synaptic weights are adjusted purely based on the correlation between pre- and postsynaptic activity \cite{hebb, palm2014}. These theories, supported by both experimental and theoretical research \cite{hebb, paulsen2000, bliss1993}, offer a biologically plausible alternative to gradient-based learning. Recent studies show that combining Hebbian plasticity with the Winner-Takes-All (WTA) mechanism \cite{hopfield19, fasthebb, grossberg1976, amato2019} can achieve comparable performance to BP in lower-layer feature learning \cite{hopfield19, fasthebb, b2}. However, the WTA mechanism is not entirely local when neurons are not interconnected, as it typically relies on global inhibition across all neurons within the same layer \cite{b2}. To address this limitation, Hebbian Principal Component Analysis (HPCA) \cite{fasthebb, oja1991, karhunen1995} introduces a mechanism of localized inhibition, restricting competition to neurons that share the same input. \cite{friston} proposed a computational model explaining how this dynamic could be realized in biological neural circuits. The model describes how dendrites strengthen synaptic connections by selectively sampling afferent inputs with highly correlated activity, reinforcing synapses that are most relevant to the network's output. At the same time, axonal arbors introduce a weakening effect by resisting the propagation of correlated signals, preventing the over-convergence of inputs that could otherwise dominate the network. This dual interaction between strengthening and weakening mechanisms ensures functional segregation, allowing for the maintenance of distinct neural pathways and minimizing excessive convergence. Such processes align with biological systems, where retrograde signaling may regulate this decay term, enhancing the biological plausibility of HPCA.

While Hebbian plasticity aligns with biological principles, it inherently lacks the discriminative learning mechanisms necessary for training the final classification layer in neural networks. As a result, bio-inspired alternatives have been developed to address this gap, such as the teacher neuron technique \cite{lagani2022, shrestha2017}, which introduces an external teacher signal to guide learning. However, this approach is biologically implausible, as it contradicts the unsupervised, local learning mechanisms of biological systems. Similarly, Contrastive Hebbian Learning (CHL) \cite{lagani2022, oreilly1996} alternates between Hebbian and anti-Hebbian phases with a supervision signal, which conflicts with the unsupervised nature of Hebbian learning. To address this, Random Contrastive Hebbian Learning (rCHL) \cite{detorakis2019} was introduced to eliminate the explicit supervision signal by introducing random feedback pathways to approximate error transmission. While this approach attempts to model brain dynamics, it relies on random feedback pathways as approximations of real brain dynamics. However, due to the incomplete understanding of biological feedback processes, these pathways may not provide an accurate or biologically plausible model \cite{detorakis2019}.

In contrast, approaches such as Forward Learning algorithms \cite{b2, hinton2022} or the method proposed by \cite{diehl} offer a simpler alternative by completely avoiding the complexities of feedback modeling. In activation learning, output activations estimate input pattern likelihoods, eliminating the need for a classification layer \cite{b2}. Weights are adjusted by enforcing competition among neurons under Hebbian principles. However, classification requires the calculation of activations for all possible classes, which is biologically implausible due to its computational inefficiency and inconsistency with the brain's energy-efficient processing. Similarly, the Forward-Forward algorithm uses the sum of squared outputs as a goodness function for training. However, it still computes activations for multiple labels, resulting in inefficiencies that make it equally implausible.

Conversely, Weight Perturbation (WP) \cite{ijcnn, bwhpc, cauwenberghs, DemboKailath} offers a biologically grounded and conceptually straightforward alternative by eliminating the need for complex feedback modeling or biologically inconsistent assumptions. WP utilizes reward-modulated amplification of random fluctuations in synaptic weights to approximate gradient directions, bypassing backward pathways and symmetric weight transport. This approach aligns naturally with mechanisms observed in the brain, offering an intuitive and biologically plausible framework for learning. However, WP faces challenges in scalability and convergence rates when applied to deeper networks, due to its reliance on unguided random search in a high-dimensional weight space \cite{review, werfel}, often referred to as the credit assignment problem. To address these limitations, we developed a bio-inspired learning rule in our earlier work, combining WP with Hebbian plasticity in the classification layer, while using only Hebbian plasticity in the hidden layers \cite{ijcnn}. Applying WP exclusively to the classification layer significantly reduces the credit assignment problem by preventing reward noise from potentially harmful upstream perturbations \cite{ijcnn}. Furthermore, this learning rule inherently induces sparsity in the weight matrices, thereby further reducing reward noise.

Building upon this foundation, the current study refines and finalizes the learning rule to enhance its performance and applicability in deep architectures. Specifically, we improve its scalability, robustness, and ability to function effectively in deeper networks and on more complex datasets. These refinements enable the rule to uniquely achieve core neurobiological properties—such as sparsity at levels compatible with cortical neural networks, decorrelation via a PCA-like mechanism, and enhanced adversarial defense \cite{wadhwa2016}. Notably, this improved formulation outperforms both standard BP and the current leading feedforward Hebbian method for image classification \cite{hopfield19}, particularly in terms of adversarial defense and generalization, especially in few-shot learning scenarios, with evaluations conducted on image data.

The sparsity introduced by our rule mitigates the credit assignment problem and aligns with biologically plausible weight distributions observed in neural systems \cite{barbour}. These properties inherently arise from the rule's design and have been recognized as crucial for efficient representations in the lower and middle layers of deep networks \cite{wadhwa2016, field1994, foldiak1990, olshausen2003, agrawal2014}. Unlike existing methods that emulate structural plasticity, such as those by \cite{knoblauch, knoblauch2017, tiddia}, Adaptive Hebbian Learning \cite{wadhwa2016}, or recent work by \cite{naresh} on Hebbian Representation Learning, which explicitly enforce these properties, our approach integrates them implicitly.

Crucially, our approach uniquely integrates these neurobiological properties without resorting to explicit enforcement, offering a solution that is more natural and aligned with biological reality. To the best of our knowledge, this is the first rule that simultaneously addresses this comprehensive set of neurobiological properties while demonstrating effectiveness on standard benchmark datasets like MNIST and CIFAR-10.

\section{Methodology}
\label{methodology}

\noindent Building on a previously developed biologically inspired learning framework \cite{ijcnn}, a set of refinements is introduced to enhance its applicability to deep neural networks. This updated framework continues to integrate competitive excitatory Hebbian plasticity \cite{ijcnn}, nonnegativity constraints, homeostatic plasticity \cite{b22}, and WP, with refinements to the weight update mechanisms and plasticity rules to ensure scalability while maintaining biological plausibility \cite{ijcnn, bwhpc}.

In the hidden layers, the original competitive Hebbian learning formulation from \cite{oja1991} is used, while maintaining the nonnegativity constraint. For the output layer, we use the adapted competitive Hebbian learning rule from previous work \cite{ijcnn} to promote stable convergence and balance between WP and Hebbian learning. Additionally, normalization methods have been integrated in the hidden layers to address the challenges of training deep networks. These modifications lead to a refined formulation of the learning rule proposed in \cite{ijcnn}.

To formally present the adjusted learning rule, the update rules for the hidden and classification layers are outlined as follows. For the hidden layers, the synaptic weight update rule is:

\begin{equation}
	\Delta w_{ij} = \eta z_j \cdot (x_i - \sum_{k} z_k w_{ik}),
\label{local}
\end{equation}

\noindent where \(x_i\) represents the input from neuron \(i\), \(\eta\) denotes the learning rate, and \(z_j = \sum_{u=1}^n x_u w_{uj}\) refers to the resulting activity of neuron \(j\).

For the classification layer, the weight updates are governed by the following combined set of rules:

\begin{equation}
	\Delta w_{ij} = \eta z_j \cdot (x_i - \sum_{k \neq j} z_k w_{ik}),
\label{classlocal}
\end{equation}

\noindent where the summation \(\sum_{k \neq j}\) excludes the \(j\)-th neuron. The WP component update rule for the classification layer is:

\begin{equation}
	\Delta w_{ki}^{\text{WP}} = - \frac{\eta}{\sigma^2} (E^{\text{pert}} - E) \xi_{ki} \mathbb{I}(w_{ki} \neq 0)
\label{classwp}
\end{equation}

\noindent where \(E^{\text{pert}}\) refers to the error of the perturbed trial, \(E\) denotes the error of the unperturbed trial, \(\xi_{ki}\) represents the perturbation term, which is a small adjustment applied to the synaptic weight to compute the perturbed output, and \(\sigma^2\) reflects the strength of the perturbation \cite{ijcnn}. The indicator function \(\mathbb{I}(w_{ki} \neq 0)\) takes the value $1$ if the synaptic weight \(w_{ki}\) is nonzero and $0$ otherwise. It is included to mitigate reward noise by ensuring that only nonzero weights contribute to the weight update \cite{ijcnn}.

In combination, both rules yield the overall learning rule for the classification layer:

\begin{equation}
	\Delta w_{ki} = \eta \alpha \cdot \Delta w^{hebbian}_{ki} + \eta \beta \cdot \Delta w^{WP}_{ki},
\label{classi}
\end{equation}

\noindent where \(\alpha\) and \(\beta\) weigh the contributions of the Hebbian and WP mechanisms, respectively.

Additionally, the bias weights are updated using:

\begin{equation}
	\Delta b_{k} = \eta \gamma \left(\frac{1}{K} \cdot \sum_{k=1}^K z_{kt} - \frac{1}{N} \cdot \sum_{t=1}^N z_{kt}\right),
\label{synscaw}
\end{equation}

\noindent where the first term denotes the mean activation of the layer, the second term represents the mean activation of the current minibatch, and \(\gamma\) controls the contribution of this learning rule \cite{b22}. For classification tasks, training batches with equally distributed labels are assumed. This simplifies the mean target activation to $\frac{1}{K}$ \cite{b22, ijcnn}.

\section{Evaluation}

\noindent This section evaluates the proposed learning rule described in section \ref{methodology}. Its performance is compared to the hebbian-based approach from \cite{hopfield19} and BP across various neural network architectures and datasets, with experiments conducted both with and without the nonnegativity constraint.

All experiments assume that input data consists of nonnegative values, a characteristic common to many real-world datasets, such as text data in bag-of-words format, pixel intensities in images, and categorical data encoded using one-hot or thermometer-scale encoding. Each input sample corresponds to a unique class label \cite{b1}.

\subsection{Experimental Setup}

\subsubsection{Studied Datasets}

\noindent Two benchmark datasets, MNIST and CIFAR-10, are used to evaluate the proposed learning rule. The MNIST dataset consists of 70,000 grayscale images of handwritten digits, each with dimensions $28 \times 28$ pixels, split into 60,000 training and 10,000 test images \cite{mnist}. MNIST is a standard benchmark for pattern recognition tasks and serves as a suitable dataset for validating biologically inspired learning rules. Its relatively low-dimensional input space and structured patterns allow us to draw parallels with biological systems that rely on localized, hierarchical processing of simple visual stimuli.

The CIFAR-10 dataset contains 60,000 RGB images of natural scenes, each with dimensions $32 \times 32$ pixels, divided into 50,000 training and 10,000 test images across 10 classes \cite{cifar10}. CIFAR-10 presents a higher degree of complexity, including variability in textures, lighting, and spatial arrangements, which align with the challenges faced by biological systems in processing naturalistic sensory inputs. Evaluating the proposed learning rule on CIFAR-10 provides insights into its ability to adapt and generalize across diverse and high-dimensional data distributions, reflecting biological mechanisms such as robustness and efficiency in representation learning.

Classification accuracy is used as the primary performance metric for both datasets, as it offers a straightforward and interpretable measure to quantify the efficacy of the proposed learning rule. For preprocessing, the pixel values of both datasets are normalized to the range $[0, 1]$, ensuring consistency in input scaling.

\subsubsection{Studied Network Architectures}
\label{architecture}

\noindent We evaluate the proposed learning rule using Multi-Layer Perceptron (MLP) architectures, primarily a model with two dense hidden layers of 2000 and 10 neurons. The input layer consists of either 784 or 3072 neurons, depending on the dataset. This setup enables a direct comparison with \cite{hopfield19}, which provides the strongest benchmark so far for bio-inspired feedforward networks, particularly on MNIST and CIFAR-10. Notably, benchmarking Hebbian learning-based feedforward networks on more complex datasets such as CIFAR-10 is an uncommon practice, making the selection of a suitable reference model crucial.

Among the available benchmarks, we prioritize the Krotov-Hopfield model \cite{hopfield19}, which stands out for its evaluation of purely excitatory networks and its elimination of the backward pass for training—a mechanism that is still not fully understood. This model is closely aligned with the core objectives of our study, as it satisfies multiple criteria for biologically plausible learning. We recognize that other benchmarks could offer additional insights, but this selection aligns most closely with the objectives of our study, which focus on addressing critical aspects of bio-inspired learning, such as performance, generalization, and robustness, within a cohesive framework.

To evaluate the scalability of the learning rule, we conduct additional experiments using deeper MLP architectures on MNIST, varying in depth with 2, 5, and 10 hidden layers. MNIST is selected for its computational efficiency and simplicity, which enable a systematic analysis of scalability without the added complexity of more challenging datasets. This complements the experiments on CIFAR-10, where scalability to more complex problems is addressed.

The stability and scalability of the proposed learning rule in deep networks, particularly under the nonnegativity constraint, necessitate a normalization procedure for both the learning rule and BP-trained networks to ensure comparability. A layer-wise training strategy is employed, where each layer computes its mean and variance to support the normalization step. The normalization function is designed to preserve activation strength, ensuring that the rescaled inputs do not disrupt the overall dynamics of the network \cite{b2}. The normalization process is defined as follows

\begin{equation}
    \text{Magnitude preserving Z-score normalization}(y) = \beta \cdot \frac{|y| - \mu_{|y|}}{\sigma_{|y|}},
   \label{normalize}
\end{equation}

\noindent where \(\mu_{|y|}\) and \(\sigma_{|y|}\) denote the mean and standard deviation of the absolute inputs \(|y|\), respectively, and \(\beta\) serves as a scaling factor to maintain the magnitude of the inputs \cite{b1}. Following this normalization, the inputs are processed through the layer using a ReLU activation function, and the outputs are subsequently rescaled to the range \([0, 1]\). This approach ensures nonnegativity while preserving activation strength across multiple layers. For networks without the nonnegativity constraint, we apply the standard ReLU activation without any specific normalization.

Weight matrices are initialized with values drawn from a uniform distribution in the range [0.01, 0.1] for nonnegativity networks, while for standard networks, they are initialized using a random normal distribution with a mean of 0.0 and a standard deviation of 0.01. Bias terms are not used in the hidden layers for any of the network architectures. For optimization, the cross-entropy loss function is employed \cite{ijcnn}.

Experiments were conducted using the Keras framework with TensorFlow as the backend. The provided benchmarks and corresponding experiments were replicated to ensure overall comparability \cite{ijcnn}.

\begin{figure}[t]
    \centering
    \begin{algorithmic}[1]
    \STATE \textbf{Initialize} weights to small, nonnegative random values.
    \FOR{each epoch}
        \STATE Shuffle training samples
        \FOR{each batch of size $N$}
            \STATE Perturb weights in classification layer
            \STATE Forward pass with \eqref{normalize} applied to hidden activations
            \STATE Save activations for Hebbian Update
            \STATE Compute cross-entropy loss
            \STATE Compute reward $(E^{\text{pert}} - E) \xi_{ki}$
            \STATE Compute $\Delta w^{WP}$ \eqref{classwp}
            \STATE Compute $\Delta w^{hebbian}$ \eqref{classlocal}
            \STATE Update $W_{output}$ \eqref{classi}
            \STATE Compute $\Delta b_{k}$ \eqref{synscaw}
            \STATE Update bias vector $B$
            \STATE Compute $\Delta w_{ij}$ for hidden layers \eqref{local}
            \STATE Update hidden layer weights $W^l$
            \STATE Set negative weights to zero (\( w = \max(0, w) \))
        \ENDFOR
    \ENDFOR
    \end{algorithmic}
	\caption{Proposed learning rule exemplified for the case of nonnegativity constraint.}
    \label{fig:spgd}
\end{figure}

\subsubsection{Training Algorithm}

\noindent The update rule for the hidden layer weights \( W^l \), where \( l \) denotes the layer index, is defined in \eqref{local}, and the output layer weights \( W_{output} \) are updated using the combined Hebbian \eqref{classlocal} and WP \eqref{classwp} rules, as shown in \eqref{classi}. The classification layer bias is updated according to \eqref{synscaw}. The training begins by initializing the weights to small nonnegative random values. In each epoch, training samples are shuffled and divided into batches of size \( N \). For each batch, a forward pass is computed, applying \eqref{normalize} to the hidden activations. The activations are saved, and the cross-entropy loss is evaluated. The loss is then compared between networks with and without weight perturbations to calculate the reward. Weight updates for the hidden and classification layers are computed using the Hebbian and WP rules, respectively. After each update, the weights and bias vectors for both layers are set to nonnegative values (\( w = \max(0, w) \)). This process is repeated for each batch and epoch.

\subsection{Results}
\label{results}

\begin{table}[t]
\caption{Classification accuracy of the proposed learning rule on MNIST and CIFAR-10 datasets, compared with benchmarks.}
\centering
\resizebox{\textwidth}{!}{%
\begin{tabular}{|c|c|c|c|c|c|c|c|c|c|}
\hline
\textbf{} & \multicolumn{4}{c|}{\textbf{ANN}} & \multicolumn{5}{c|}{\textbf{ANN with Nonnegativity Constraint}} \\
\cline{2-10} 
\textbf{Accuracy} & \multicolumn{3}{c|}{MNIST} & CIFAR10 & \multicolumn{4}{c|}{MNIST} & CIFAR10 \\
\cline{2-10}
& 1 Layer$^{\mathrm{a}}$  & 1 Layer$^{\mathrm{b}}$ &  2 Layers  & 1 Layer & 1 Layer$^{\mathrm{a}}$ & 1 Layer$^{\mathrm{b}}$ & 2 Layers & 10 Layers & 1 Layer \\
\hline
Authors$^{\mathrm{d}}$ & $97.45\%$ & $95.46\%$ & $97.34\%$ & $46.07\%$ & $94.20\%$ & $90.77\%$ & $91.82\%$ & $91.33\%$ & $33.98\%$ \\
\hline
K\&H \cite{hopfield19} & $98.54\%$ & $92.56\%$ & $88.31\%$ & $55.05\% ^{\mathrm{c}}$ & $97.98\%$ & $90.31\%$ & $30.02\%$ & N/A & N/A \\
\hline
BP & $98.26\%$ & N/A & $98.53\%$ & $55.27\%$ & $94.47\%$ & N/A & $93.82\%$ & $11.35\%$ & $36.28\%$ \\
\hline
\end{tabular}%
}
\label{res}
\begin{flushleft}
$^{\mathrm{a}}$ BP was used to train the output layer in this column for comparability.
\end{flushleft}
\begin{flushleft}
$^{\mathrm{b}}$ These columns show results from extended training with the proposed learning rule and \cite{hopfield19}, which uses their rule with ReLU activation.
\end{flushleft}
\begin{flushleft}
$^{\mathrm{c}}$ The results for CIFAR-10 in \cite{hopfield19} could not be replicated due to the absence of detailed hyperparameter specifications provided in the original work.
\end{flushleft}
\begin{flushleft}
$^{\mathrm{d}}$ Hyperparameters for the proposed method across all experiments: $\eta = 0.0005$, $\sigma^2 = 0.00016$, $\alpha = 0.04$, $\beta = 87500.00$, $\gamma = 0.04$. For WP-based output layer experiments, a batch size of 2000 was used.
\end{flushleft}
\end{table}

\begin{figure*}[b] 
    \centering
    \includegraphics[width=\textwidth]{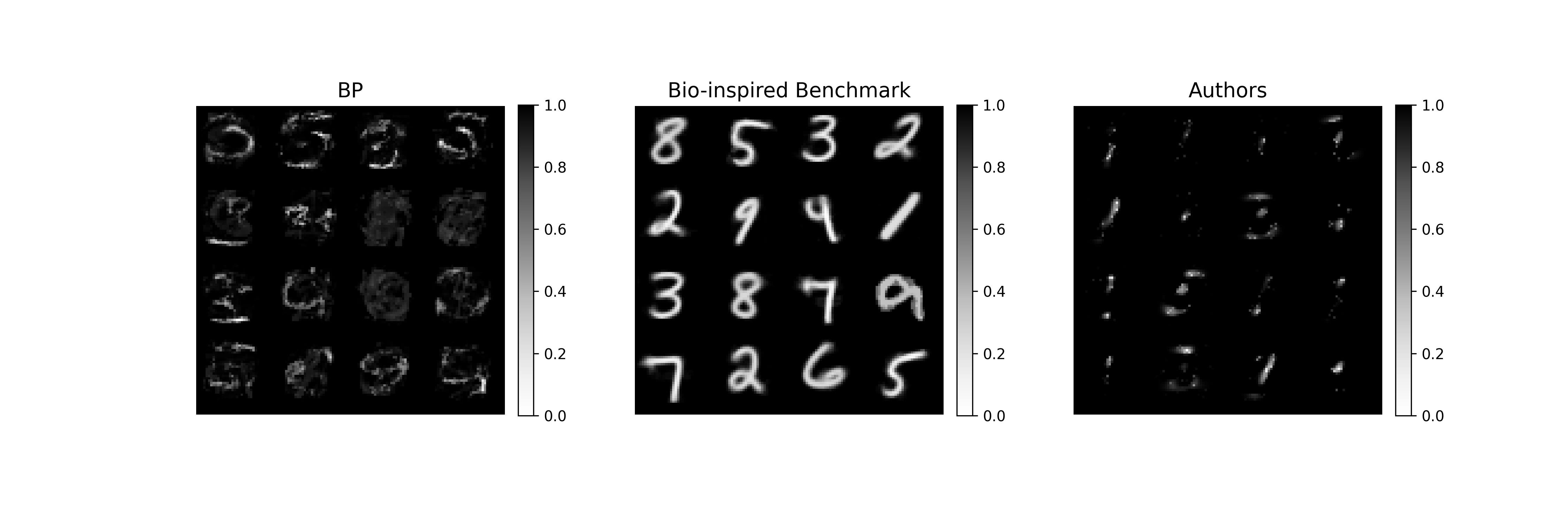} 
    \caption{Weight kernels for networks trained with the proposed learning rule, the method of \cite{hopfield19}, and BP, all incorporating the nonnegativity constraint.}
    \label{fig:example2} 
\end{figure*}

\noindent Table~\ref{res} summarizes the results of our experiments, showing the classification accuracy on the test sets for different network topologies, datasets, and learning rules. The performance is compared to the baseline BP and the method proposed by \cite{hopfield19}. This table highlights the relative effectiveness of the developed learning rule in various configurations, offering insights into how it performs compared to other established methods.

The proposed learning rule initially demonstrates slightly lower performance when compared to the benchmarks utilizing the same 1-hidden-layer network architectures as in \cite{hopfield19} (see Table~\ref{res}). This discrepancy can be attributed to BP's ability to compute the optimal gradient, whereas the WP component of the proposed learning rule approximates this gradient, ultimately converging to a terminal error \cite{ijcnn}. However, even when integrating the proposed learning rule with a BP-trained output layer, a minor performance gap persists.

This gap can be attributed to the fact that correlation-based learning rules, such as Hebbian plasticity, are unable to strengthen synapses at the borders of digits, where discriminative power is higher compared to the central pixels shared across multiple classes. BP, on the other hand, enhances these connections, improving class discrimination (see Fig.~\ref{fig:example2}) \cite{ijcnn}. Networks trained by the learning rule from \cite{hopfield19}, in contrast, store complete images, memorizing all features. The proposed learning rule, however, encodes partial patterns, capturing stable central features that are shared across multiple variations of digits.

Both the proposed learning rule and BP experience a notable reduction in performance when the nonnegativity constraint is applied. This finding is consistent with prior research \cite{b1}, which demonstrated that nonnegativity negatively impacts classification accuracy in BP-trained feedforward networks, providing a clear rationale for the observed decline.

Nevertheless, the performance of the proposed learning rule ranks at the higher end of the range formed by the accuracies reported by other bio-inspired algorithms in the literature, such as those presented by \cite{scellier, sacramento, lee, diehl}, with accuracies ranging from $95.0\%$ to $98.06\%$. However, the core objective of this work is not solely focused on maximizing performance, but rather on achieving competitive performance while simultaneously ensuring a comprehensive biologically plausible implementation at an abstracted level. Therefore, we focus our detailed comparisons primarily on \cite{hopfield19}, as discussed in \ref{architecture}. This decision is based on the fact that all of the aforementioned algorithms either face the issue of modeling backward feedback connections with uncertain or inadequate biological plausibility, as detailed in \ref{introduction}, or exhibit lower accuracy, as observed in the work by \cite{diehl}. As a result, not all of these benchmarks are included in Table~\ref{res}, as our detailed comparison targets those that most closely align with the objectives and scope of this work.

Finally, the slower convergence behavior of the WP component in the proposed learning rule necessitated extended training until saturation, which yielded a 2.5\% improvement in performance. As a result, the performance of the proposed learning rule was approximately 3.7\% below that achieved by a BP-trained network with the same hidden layer architecture.

\subsubsection{MNIST}

\noindent \cite{hopfield19} have demonstrated performance that can even surpass standard BP \footnote{The authors note that the performance of networks trained end-to-end with BP can be significantly enhanced using techniques such as dropout, noise injection, and data augmentation.}. However, this outperformance of BP can be attributed to their manual tuning of hyperparameters for their power law activation functions, which are adjusted layer-wise. In contrast, when utilizing standard activation functions like ReLU, performance is reduced to $92.56\%$. The proposed learning rule , however, outperforms this result, regardless of whether biologically inspired or BP-trained classification layers are used. Notably, the application of nonnegativity constraints to networks trained with the method from \cite{hopfield19} further decreases accuracy leading to a test accuracy of $90.31\%$, which aligns with the performance of networks trained with the proposed learning rule using the WP output layer.

Increasing network depth leads to significant performance degradation for the method from \cite{hopfield19}, especially for architectures with the nonnegativity constraint. While the proposed learning rule also experiences performance decreases with increased depth, these are marginal and less pronounced compared to \cite{hopfield19}'s approach. In contrast, BP shows minimal sensitivity to changes in network depth, with performance varying by around $0.2-0.5\%$. However, when the nonnegativity constraint and normalization process are applied, BP performance significantly decreases for deeper architectures, achieving only 11.35\% accuracy with a 10-hidden-layer network. In contrast, the proposed learning rule exhibits only minor performance reductions, with accuracy decreasing by up to 3\% for the deepest architecture.

\subsubsection{CIFAR-10}

\noindent For CIFAR-10, all considered learning rules exhibited subpar performance, consistent with prior findings for feedforward networks on this dataset. While CIFAR-10 is generally considered more challenging than MNIST, the observed trends align with those seen on simpler datasets, with CIFAR-10 serving primarily as an additional benchmark. The proposed learning rule yields lower performance on CIFAR-10 compared to \cite{hopfield19} (see Table~\ref{res} for details). Part of this performance gap may be attributed to overfitting in their method, where activation functions are manually tuned and adjusted layer-wise. Nevertheless, the performance of the proposed learning rule remains consistent with other bio-inspired algorithms in the literature \cite{bartunov2018, hopfield19}, where accuracies range between $40.86\%$ and $58.03\%$.

\subsubsection{Synaptic Weights Distribution for MNIST}

\noindent In this subsection, we focus on the synaptic weight distributions of networks trained with BP, our learning rule, and the approach from \cite{hopfield19}, specifically under the nonnegativity constraint. These distributions were evaluated and compared to lognormal, as well as compressed and stretched exponential distributions, which have been proposed in the neuroscience literature as models for synaptic weight distributions in biological neural networks \cite{barbour, brunel, song2005, varshney, teramae2014}. Therefore, a close alignment of the observed weight distributions with lognormal or compressed exponential forms strongly supports the enhanced biological plausibility of the model, suggesting that the network's behavior mirrors key characteristics observed in biological neural systems.

\begin{figure*}[t]
    \centering
    \includegraphics[width=\textwidth]{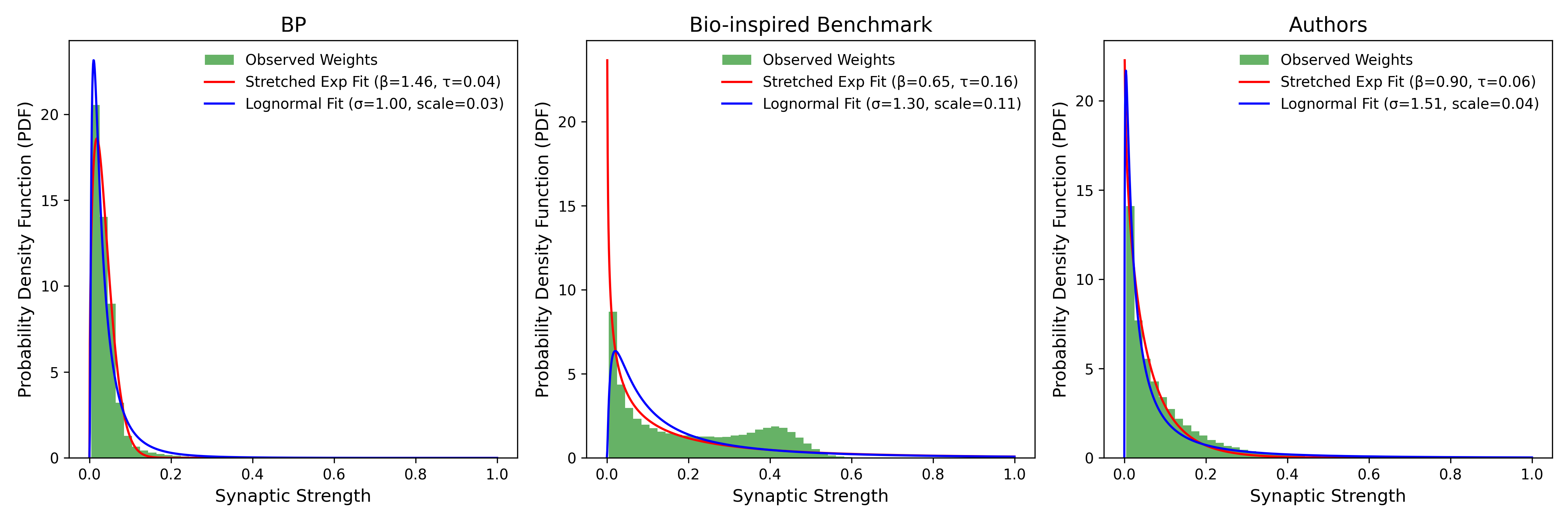} 
    \caption{Comparison of synaptic weight distributions for networks trained with BP, the proposed learning rule, and \cite{hopfield19} under a nonnegativity constraint, fitted to lognormal, compressed, and stretched exponential distributions. Synaptic weights were normalized between 0 and 1 and clipped at 0.005 to account for the synaptic detection threshold. Tests showed that weights below this threshold had no significant impact on performance, while higher thresholds (e.g., 0.01) distorted the distributions. The chosen threshold of 0.005 preserved relevant synaptic connections without distorting the distribution.}
    \label{fig:example3} 
\end{figure*}

The synaptic weight distributions for all approaches exhibit sparsity. However, the degree of sparsity differs notably between methods. Networks trained with our learning rule and BP show approximately 90\% of weights being zero, indicating a high degree of sparsity, compatible with cortical neural networks \cite{knoblauch, braitenberg1991}. In contrast, networks trained with \cite{hopfield19}'s approach result in a significantly lower sparsity of around 67\%.

Both the lognormal and exponential distributions closely resemble each other in terms of curve shape for the synaptic weight distributions observed in \cite{hopfield19} and the proposed learning rule. For our learning rule, both distributions provide an exceptionally close fit to the observed synaptic weight distribution. In the case of BP, the lognormal distribution fits closely, while the exponential distribution fails to capture the right tail. In contrast, Networks trained by \cite{hopfield19} exhibit a bimodal distribution, the most divergent from the expected biological distribution. Neither the lognormal nor the exponential distribution is bimodal, highlighting the largest discrepancy with biological reality among the methods evaluated.

\subsubsection{Adversarial Robustness and Generalization}
\label{general}

\noindent This subsection examines the adversarial robustness and generalization capabilities of networks trained using BP, our learning rule, and the method in \cite{hopfield19}, under the nonnegativity constraint. Adversarial robustness is particularly relevant in evaluating the biological plausibility of the proposed learning rule, as biological neural networks exhibit inherent resilience to adversarial perturbations \cite{zador2019, hennequin2014}. For robustness evaluation, we apply adversarial perturbations generated via the Fast Gradient Sign Method (FGSM) and Projected Gradient Descent (PGD) attacks, which are widely used benchmarks for assessing the vulnerability of deep networks \cite{goodfellow2015, madry2018}. We measure classification accuracy under varying perturbation magnitudes, with a smaller accuracy drop indicating improved robustness in adversarial settings.

\begin{figure*}[t]
    \centering
    \includegraphics[width=\textwidth]{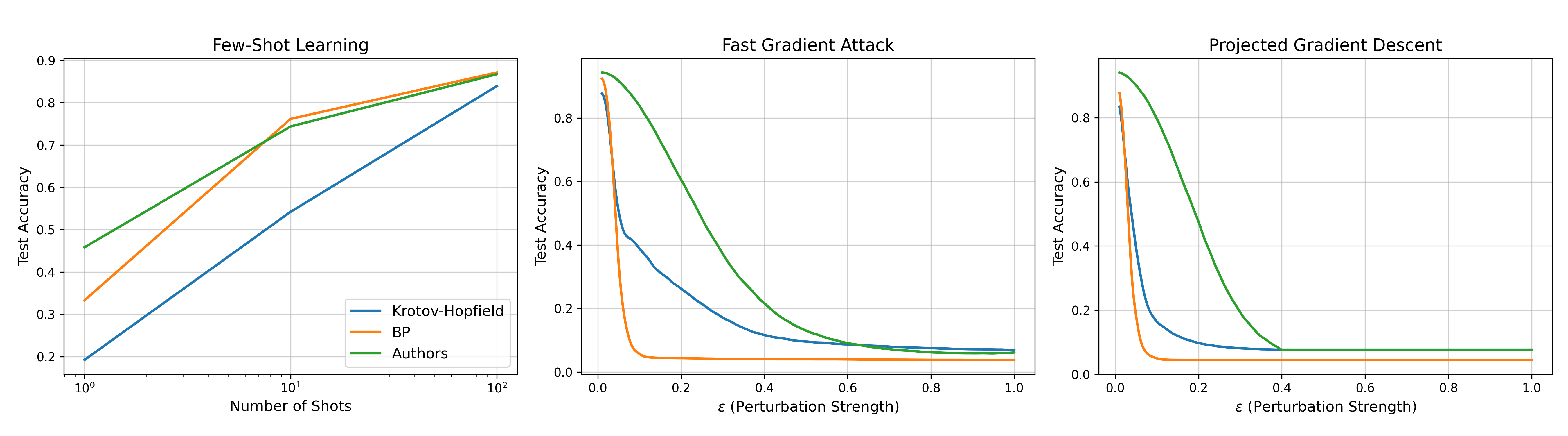} 
 \caption{Comparison of test accuracy for networks trained with BP, the proposed learning rule, and \cite{hopfield19} under different conditions: Few-Shot Learning, Fast Gradient Attack, and Projected Gradient Descent. The x-axis represents the number of shots (left) and perturbation magnitude $\epsilon$ (middle, right). For the PGD attack, we used a step size of 0.01, a total of 40 iterations, and an $\ell_{\infty}$ norm constraint.
}
    \label{adverse} 
\end{figure*}

As shown in Fig.~\ref{adverse}, accuracy declines across all approaches as the perturbation magnitude increases, which is expected. Specifically, the PGD attack results in a more severe drop compared to the FGSM attack, as PGD is an iterative method that refines perturbations over multiple steps, making it harder to defend against. BP exhibits a sharp accuracy decline under both attack types, whereas bio-inspired approaches show a more gradual decline under FGSM. However, under PGD, \cite{hopfield19} shows a sharp drop, indicating its vulnerability to stronger adversarial attacks.

In contrast, our learning rule demonstrates superior robustness, maintaining a stable performance under PGD up to a perturbation magnitude of 0.1, beyond which accuracy declines linearly. This suggests that the learned representations are more resilient to adversarial modifications. Such behavior aligns with the observation that biological neural networks exhibit inherent robustness to perturbations due to their distributed and redundant encoding mechanisms \cite{zador2019, hennequin2014}.

In addition to robustness, we also evaluate the generalization capabilities of the model in few-shot learning settings, specifically assessing 1-shot, 10-shot, and 100-shot scenarios. This evaluation is motivated by the observation that biological neural networks generalize effectively from sparse supervision \cite{zador2019, lake2017}. Few-shot learning is commonly used to assess a model's ability to generalize from limited data. Improved performance in few-shot learning suggests that the learned representations exhibit greater efficiency.

Performance declines as the number of labeled examples decreases, but the extent of this decline varies across methods. \cite{hopfield19} struggles in the 1-shot setting, achieving only 10\%-20\% accuracy with minimal improvement over performance before training. In contrast, the proposed learning rule consistently outperforms BP and \cite{hopfield19} in 1-shot learning, achieving 45\%-55\% accuracy depending on the subset. While it does not yet match the overall accuracy of state-of-the-art methods such as convolutional Siamese networks or Hierarchical Bayesian Program Learning, its performance is comparable to that of feedforward Siamese networks, while strictly adhering to neurobiological constraints \cite{siamese}.

\section{Discussion}

\noindent In this paper, we demonstrate that the proposed learning rule inherently achieves several fundamental neurobiological properties. Specifically, the model inherently satisfies Dale's Law, as it is designed for excitatory networks, where the nature of the connections ensures compliance by definition. The experimental results further reveal biologically plausible sparsity, with a connectivity rate of approximately 10\%, and a lognormal distribution of weights. Furthermore, the enhanced model demonstrates robust adversarial defense. When compared to the leading bio-inspired feedforward model for MNIST and CIFAR proposed by \cite{hopfield19}, our approach exhibits superior generalization, particularly in few-shot learning scenarios. These results emphasize the holistic nature of the proposed method, wherein the neurobiological properties emerge implicitly, without reliance on explicitly defined parameters, distinguishing it from alternative approaches.

Despite these advantages, when comparing classification accuracy to the bio-inspired approach of \cite{hopfield19}, we observe some notable differences. Specifically, their approach outperforms both our model and BP in single-hidden-layer ANNs. This performance gap arises from the use of biologically implausible activation functions, in particular extreme power-law functions. While optimized for specific tasks, these functions are inconsistent with natural neural dynamics, where neurons typically exhibit more gradual and bounded responses. Using activation functions such as ReLU, that are closer to these neural dynamics on an abstract level, their accuracy substantially decreases. In contrast, the proposed learning rule achieves superior performance when utilizing BP-trained output layers.

Moreover, the reliance on extreme exponents in the method of \cite{hopfield19} introduces substantial challenges, particularly when scaling to deeper architectures. These extreme exponents can lead to uncontrolled activations, disrupting the stability of signal propagation and impairing the network's ability to effectively learn and generalize. This issue becomes evident in Table~\ref{res}, where introducing a single additional hidden layer to their approach results in notable performance degradation, especially under the nonnegativity constraint. In contrast to these limitations, the proposed learning rule scales effectively to deeper architectures. This is particularly significant, as deeper networks generally lead to LTP due to the interplay between Hebbian plasticity and nonnegativity. As demonstrated in Table~\ref{res}, the proposed learning rule maintains stability and adaptability even within complex networks of up to 10 hidden layers. While performance is effectively sustained, only minor degradation occurs due to the convergence behavior of the proposed learning rule. Since weight updates rely on local information, the number of epochs needed to reach optimal performance increases with network depth. Each layer's ability to converge depends on the proximity of the preceding layer to its optimum, creating a sequential dependency. Consequently, convergence slows in deeper layers. However, the slower convergence observed here is primarily a theoretical challenge rather than a practical limitation. In biological systems, such as the visual cortex, complex information is processed efficiently with a relatively limited number of layers \cite{felleman1991, zeki2017}.

In stark contrast, BP faces significant challenges as network depth increases, particularly in excitatory architectures. As the number of layers increases, BP's performance deteriorates due to vanishing gradient effects, as demonstrated in Table~\ref{res}. Unlike the proposed learning rule, BP lacks inherent mechanisms such as homeostatic or synaptic scaling to mitigate this issue.

In shallow networks, BP marginally outperforms the proposed learning rule when using a BP-trained output layer, achieving a performance advantage of 0.27\%. As shown in \ref{results}, BP strengthens connections at digit borders, while the proposed learning rule emphasizes stable, shared features across digit variations, which accounts for this minor performance gap. This approach reduces overfitting by retaining only the most critical features, which becomes particularly evident when comparing 1-shot performance in Fig.~\ref{adverse}. This behavior aligns with the hypothesis that biological systems can effectively extract structure from sparse input, likely leveraging local learning mechanisms and distributed representations \cite{lake2017}. Such behavior suggests that our model is able to emulate key aspects of biological learning.

Furthermore, this partial encoding enhances the network's resilience to adversarial perturbations. Unlike networks trained by \cite{hopfield19}, which memorize entire patterns, the proposed learning rule encourages a more generalized representation, aligning with the distributed encoding strategies of biological neural networks \cite{zador2019, hennequin2014}, as shown in Fig.~\ref{fig:example2}. While our rule outperforms both \cite{hopfield19} and BP in adversarial defense, we recognize that other bio-inspired mechanisms, such as dendritic saturation \cite{nayebi2017} and RAILS \cite{wang2017}, provide superior state-of-the-art defense capabilities. However, these methods tend to prioritize adversarial robustness at the expense of broader neurobiological processes. For example, \cite{nayebi2017} focuses on activation functions based on dendritic saturation, but still relies on conventional neural networks and gradient-based optimization, while RAILS neglects important neurobiological properties like sparsity and locality in learning \cite{wang2017}.

The principles behind the partial encoding in our model are rooted in the well-established framework of PCA \cite{oja1991}, where the learning rule naturally captures key features of the input data \cite{oja1991}. In the absence of normalization constraints, the model simplifies to the famous Oja's rule for Neural PCA \cite{oja1991}, where neurons form stimulus-selective receptive fields that align with the principal components of the input data. With the incorporation of nonlinear activation functions, as in our model, the network can also capture higher-order correlations, such as independent components \cite{oja1991, ocker2021}. This behavior mirrors the selective processes observed in biological systems, such as simple cells in the visual cortex. These processes have been shown to produce similar feedforward receptive fields \cite{bell1997, brito2016}, further reinforcing the biological plausibility of our approach \cite{eckmann2024}.

In particular, \cite{oja1991} demonstrated that a similar rule enables the weight matrix to effectively perform independent component analysis (ICA). Given that the proposed learning rule follows the same update process with nonlinear activation, it can be considered an instance of Oja's framework, allowing us to leverage the insights from his analysis. Upon convergence, the columns of the trained weight matrix represent estimates of the basis vectors of ICA, in any order. This enables the reconstruction of input data through the output of the trained layer. This ability enhances the resilience of the learning process, particularly against adversarial perturbations, as can be seen in Fig.~\ref{adverse} \cite{oja1991, qin}.

However, biological systems must balance efficiency and robustness, due to space and energy limitations in the human brain \cite{knoblauch, tiddia}. This balance is also a challenge for Hebbian learning methods, which encounter significant scaling difficulties when applied to larger datasets and architectures \cite{fasthebb}. Recent advancements, such as FastHebb \cite{fasthebb}, have addressed these challenges by merging update computation with batch aggregation and leveraging GPU-optimized matrix multiplication techniques. These optimizations result in up to 50-fold improvements in training speed, enabling, for the first time, the application of Hebbian learning algorithms to large-scale datasets like ImageNet and deep architectures such as VGG. Such developments demonstrate the feasibility of scaling Hebbian learning to more complex scenarios while retaining its biologically inspired foundation.

In this context, convolutional neural networks (CNNs) offer a promising architecture for further exploration, aligning closely with the brain's visual processing mechanisms. Recent research shows that applying Hebbian plasticity to CNNs can improve performance by up to 40\% on tasks like CIFAR-10 compared to training feedforward networks with Hebbian learning \cite{fasthebb, lagani2022}. Future research will focus on exploring these advantages.

However, even when hidden layers are replaced with convolutional layers, challenges remain with the WP component of the proposed learning rule. While WP demonstrates biological plausibility and competitive performance, it is computationally demanding. WP-trained layers suffer from the unguided nature of random weight searches in high-dimensional spaces \cite{review, werfel}. Despite the mitigation strategies incorporated into the proposed learning rule, achieving BP-comparable performance requires significantly more epochs. Under the same training duration as the experiments of \cite{hopfield19}, the WP-trained model achieves only 88\% accuracy, which slightly underperforms.

Although the WP-trained classification layer requires more epochs to achieve BP-comparable performance, this slower convergence mirrors the self-regulating dynamics of biological systems, where the model autonomously reaches an equilibrium state, without relying on manual interventions like epoch tuning or early stopping to avoid overfitting. Unlike BP and the approach in \cite{hopfield19}, which depend on such manual adjustments, the proposed rule inherently avoids these challenges through its intrinsic self-regulation. The WP component is driven by reward, which is linked to the distance of weights from their optimal values, gradually decreasing as training progresses. Concurrently, Hebbian plasticity is guided by stable activity patterns and the inherent characteristics of the dataset, stabilizing the model and eliminating the need for external interventions.

While this stable asymptotic convergence leads to terminal error for finite training durations, it prevents the network from memorizing all of the training data, as observed in BP and \cite{hopfield19} networks. Without external interventions, such models utilize all available resources, leading to overfitting. This behavior contrasts with the brain's efficient storage mechanisms, which are constrained by both space and energy limitations \cite{knoblauch, tiddia}. 

In biological neural networks, this efficiency is evident in the distribution of synaptic weights. Research has shown that synaptic weights in the cortex follow lognormal distributions, which are consistent with the natural constraints of biological systems \cite{song2005, teramae2014}. Furthermore, studies of excitatory postsynaptic potentials (EPSPs) reveal similar distribution patterns, providing insights into the underlying distribution of synaptic weights. These findings suggest that biological networks prioritize efficient resource allocation, avoiding redundancy while maximizing storage capacity.

Accurately modeling these synaptic distributions is challenging due to zero and near-zero weights masked by the synaptic detection threshold \cite{varshney}. While lognormal distributions are commonly observed, alternative models, such as truncated Gaussian distributions \cite{barbour} and stretched exponential distributions \cite{varshney}, have been fitted to various biological datasets, offering a closer alignment to the observed synaptic distributions. However, as shown in Fig.~\ref{fig:example3} and in \cite{song2005}, these fitted distributions often closely resemble each other, exhibiting a similar pattern of monotonically decaying from zero weight \cite{barbour, varshney}. Specifically, \cite{barbour, brunel} demonstrates that a distribution fitting this pattern is observed in the synapses between cerebellar granule cells and Purkinje cells \footnote{Purkinje cells are neurons in the cerebellar cortex that play a crucial role in motor control by integrating extensive sensory and motor information and releasing the inhibitory neurotransmitter GABA, which modulates nerve impulse transmission and coordinates motor movements.}, where nonnegative weight constraints maximize storage capacity while simultaneously accommodating silent synapses. This behavior not only enhances storage efficiency but also improves classification performance, reflecting the brain's optimization of resource utilization under structural, developmental, and energy constraints \cite{barbour, zhong2022theory}.

The proposed learning rule generates weight distributions that closely resemble those found in biological systems, while also aligning with those produced by gradient-based optimization in excitatory perceptrons when constrained by biologically inspired mechanisms. This suggests that, under these constraints, Hebbian plasticity can approximate the mathematically optimal weight distributions generated by constrained BP \cite{barbour, brunel}. Furthermore, the sparsity enforced by the proposed learning rule mirrors the efficient resource utilization seen in biological neural networks, where only a small fraction of synapses are active at any given time \cite{barbour}. By employing local competition and nonnegativity constraints, the proposed learning rule aligns synaptic weights with the covariance structure of the input data, mirroring the brain's efficient resource allocation. This enables neurons within the same layer to represent different features, adhering to the constraints of nonnegative weight distributions and efficient resource allocation.

In contrast to traditional methods for emulating structural plasticity \cite{knoblauch, tiddia}, which rely on predefined constraints or manual optimization, the proposed learning rule dynamically adapts synaptic and neural structures based on the data's characteristics \cite{bwhpc}. This dynamic adjustment ensures efficient resource allocation by prioritizing task-relevant features while minimizing unnecessary complexity. Additionally, the proposed learning rule supports ongoing synaptic rewiring, a key aspect of structural plasticity in biological systems \cite{knoblauch, tiddia}, enabling continuous optimization without the need for external intervention.

This capability is particularly evident in our preliminary results, where we trained a single model simultaneously on both the MNIST and Urban8K sound datasets. Through this concurrent training approach, we observed a combined accuracy comparable to single-task training, even without explicitly designating task-specific neuronal allocations. While these findings are promising, they are preliminary and require further research. They suggest that the model could autonomously manage multiple tasks by encoding task-specific sub-features across diverse data modalities. This potential for multimodal learning is intriguing and may offer insights into how the human brain allocates neural resources for task-specific processes using unsupervised, activity-driven plasticity. Further investigation is underway to rigorously test this hypothesis and explore its broader applications.

\section{Conclusion}

\noindent This study advances the application of biologically plausible principles in neural networks. The proposed rule induces sparsity while adhering to further core neurobiological principles, including decorrelation, Dale's law, lognormal synaptic weight distributions, and natural convergence, without explicit enforcement. These principles guide network behavior that mimics biological systems, which in turn enhances robustness against adversarial attacks and improves generalization, particularly in few-shot learning scenarios. This behavior emerges naturally from the model's alignment with neurobiological constraints, positioning our approach as a strong candidate for future applications in biologically inspired neural networks.

While the proposed method shows promising results, further refinement is needed to optimize its performance on classification tasks. A primary challenge is the WP component in the output layer, which, despite previous mitigation efforts, remains a computational bottleneck. Additionally, the reliance on Hebbian learning across both the output and hidden layers limits scalability, particularly when applied to larger network architectures and more complex image datasets. To address these limitations, integrating FastHebb \cite{fasthebb} is essential. Furthermore, exploring CNNs is necessary for effectively scaling the method to larger, more complex datasets. FastHebb has already demonstrated superior scalability in such settings, making it critical for adapting the learning rule to CNNs. In addition, exploring spiking neurons and neuromorphic hardware offers a promising approach to improving computational efficiency, as they provide a natural framework for temporal information processing and enable event-driven computation \cite{review, maass1997, renner2021, yang2023}.

\end{document}